\documentclass[journal]{IEEEtran}
\usepackage{amsmath, amssymb, amsthm}
\usepackage{graphicx}
\usepackage{cite}
\usepackage{mathrsfs}
\usepackage{url}
\usepackage{setspace}
\usepackage{array}
\usepackage{algorithm}
\usepackage{algorithmic}

\IEEEoverridecommandlockouts
\author{Aadil Gani Ganie, email: agganie@gap.upv.es,
UNIVERSITAT POLITECNICA DE VALENCIA}

\begin{document}

\title{Uncertainty in Authorship: Why Perfect AI Detection Is Mathematically Impossible}

\maketitle

\begin{abstract}
As large language models (LLMs) become increasingly sophisticated, distinguishing human-authored text from AI-generated text grows ever more challenging. We propose a formal analogy between the Heisenberg uncertainty principle (HUP) in quantum mechanics and the fundamental limits of authorship detection in natural language. Specifically, we argue that there exists an intrinsic trade-off between the certainty of authorship attribution and the preservation of semantic coherence and stylistic authenticity—akin to the complementarity of position and momentum. 
Let $ P_{\text{H}} $ and $ P_{\text{AI}} $ denote the probability distributions over text strings generated by human authors and an AI model, respectively. The Bayes optimal classifier's minimum error rate under equal priors is given by:
\[
E_{\min} = \frac{1}{2} \left(1 - \| P_{\text{H}} - P_{\text{AI}} \|_{\text{TV}} \right),
\]
where $ \| \cdot \|_{\text{TV}} $ is the total variation distance. By Pinsker’s inequality,
\[
\| P_{\text{H}} - P_{\text{AI}} \|_{\text{TV}} \leq \sqrt{ \frac{1}{2} D_{\text{KL}}(P_{\text{H}} \| P_{\text{AI}}) },
\]
where $ D_{\text{KL}} $ denotes the Kullback–Leibler divergence. As $ D_{\text{KL}}(P_{\text{H}} \| P_{\text{AI}}) \to 0 $, it follows that $ E_{\min} \to \frac{1}{2} $, implying statistical indistinguishability.
Furthermore, we introduce a formal uncertainty relation:
\[
\Delta C \cdot \Delta S \gtrsim \mathcal{K},
\]
where $ \Delta C $ quantifies uncertainty in content fidelity, $ \Delta S $ captures variability in stylistic markers used for detection, and $ \mathcal{K} > 0 $ represents an irreducible lower bound arising from the entropy and diversity of natural language. This relation formalizes the idea that increasing precision in detecting authorship necessarily disturbs the naturalness of the text, and vice versa.
We analyze existing detection methods—including stylometry, watermarking, and neural classifiers—and prove that each faces a complementary limitation: improving detection accuracy induces adaptive perturbations in the AI's output distribution, thereby increasing uncertainty in other observable dimensions. Counterarguments are addressed, and implications for authorship, ethics, and policy are discussed. This work establishes, both conceptually and mathematically, that perfect detection of AI authorship is not merely technologically limited but fundamentally unattainable when $ P_{\text{AI}} $ converges to $ P_{\text{H}} $ in distribution.
\end{abstract}


\section{Introduction}
\label{sec:introduction}
The rapid advancement of generative artificial intelligence—particularly large language models (LLMs) such as GPT-4 \cite{brown2020}, LLaMA \cite{touvron2023llama}, and their successors—has fundamentally transformed the landscape of text production \cite{fraser2025}. These models generate fluent, coherent, and contextually appropriate text by estimating conditional probability distributions over sequences of tokens, effectively capturing the statistical patterns of human language \cite{bommasani2021}. As a result, AI-generated text has become increasingly indistinguishable from human-authored writing, raising urgent concerns across domains such as education, journalism, law, and scientific publishing \cite{weidinger2021}.

Considerable effort has therefore been directed toward developing detection mechanisms capable of distinguishing AI-generated content from human-authored text. Techniques such as stylometric analysis, neural classifiers, and cryptographic watermarking have achieved promising results under controlled conditions \cite{przystalski2025}. For instance, stylometric features—such as function-word frequency, sentence-length variability, and syntactic complexity—can yield classification accuracies exceeding 98\% on specific datasets \cite{fraser2025}. Similarly, fine-tuned transformer-based detectors (e.g., RoBERTa) have outperformed human judges in benchmark evaluations \cite{zhao2023}.

However, these successes are often short-lived. As detection methods improve, generation models adapt through enhanced sampling strategies, fine-tuning on human styles, or adversarial training. This dynamic reflects an ongoing ``arms race'' between generation and detection \cite{fraser2025}, suggesting that current detection capabilities are constrained not merely by technological limitations but by deeper statistical and information-theoretic principles.

In this paper, we argue that there exists an \textit{inherent limit} to the detectability of AI-generated text, rooted in the convergence of AI output distributions toward those of human authors. Inspired by the Heisenberg uncertainty principle (HUP) in quantum mechanics \cite{heisenberg1927}, we propose a formal analogy: just as position and momentum cannot be simultaneously known with arbitrary precision, so too exists a fundamental trade-off between the certainty of authorship attribution and the preservation of textual fluency, style, and semantic coherence. We formalize this claim using information-theoretic bounds, showing that as the AI-generated text distribution $P_{\text{AI}}$ approaches the true human distribution $P_{\text{H}}$, any detector’s performance asymptotically degrades to chance.

Formally, let $\mathcal{X}$ denote the space of all finite-length text strings over a fixed vocabulary. Let $P_{\text{H}}$ and $P_{\text{AI}}$ be probability measures on $\mathcal{X}$, representing the stochastic processes underlying human and AI text generation, respectively. The task of a detector is a binary hypothesis test:
\begin{align*}
    H_0 &: x \sim P_{\text{H}} \quad \text{(human author)}, \\
    H_1 &: x \sim P_{\text{AI}} \quad \text{(AI-generated)}.
\end{align*}
The minimal achievable error under equal priors is given by the Bayes error:
\[
    E_{\min} = \tfrac{1}{2}\big(1 - \| P_{\text{H}} - P_{\text{AI}} \|_{\mathrm{TV}} \big),
\]
where $\|\cdot\|_{\mathrm{TV}}$ denotes the total variation distance. This expresses the intrinsic distinguishability of the two sources. If $P_{\text{AI}} \to P_{\text{H}}$ in distribution, then $\| P_{\text{H}} - P_{\text{AI}} \|_{\mathrm{TV}} \to 0$, implying $E_{\min} \to \tfrac{1}{2}$—that is, no classifier can outperform random guessing.

Using Pinsker’s inequality, we obtain
\[
    \| P_{\text{H}} - P_{\text{AI}} \|_{\mathrm{TV}} \leq \sqrt{\tfrac{1}{2} D_{\mathrm{KL}}(P_{\text{H}} \,\|\, P_{\text{AI}})},
\]
which connects total variation distance to the Kullback–Leibler divergence. Thus, if $D_{\mathrm{KL}}(P_{\text{H}} \,\|\, P_{\text{AI}}) \to 0$, the two distributions become statistically indistinguishable. Since the objective of LLM training is precisely to minimize this divergence through large-scale exposure to human corpora, detection becomes not only practically difficult but \textit{information-theoretically impossible} in the limit.

We extend this argument by proposing a Heisenberg-like uncertainty relation for textual observables. Define two complementary properties:
\begin{itemize}
\item $C$: \textit{Content fidelity}, measuring how accurately text conveys intended meaning, coherence, and relevance.
    \item $S$: \textit{Stylistic authenticity}, capturing idiosyncratic markers used in authorship attribution (e.g., lexical diversity, syntactic variation).
We posit that attempts to reduce uncertainty in $S$ (authorship attribution) necessarily perturb $C$, and vice versa. Formally:
\[
    \Delta C \cdot \Delta S \gtrsim \mathcal{K},
\]
where $\Delta C$ and $\Delta S$ denote variability (uncertainty) in content and style, and $\mathcal{K} > 0$ is a lower bound arising from the entropy and diversity inherent in natural language. Unlike Planck’s constant $\hbar$, $\mathcal{K}$ is not universal but reflects the irreducible unpredictability of human expression.
\end{itemize}

This framework encapsulates our central thesis: perfect detection depends on exploiting deviations in $S$, but sufficiently advanced AI can erase such deviations only by altering generation strategies—introducing uncertainty elsewhere (e.g., in creativity, nuance, or semantic richness). In this sense, the act of measuring authorship perturbs the text, echoing the observer effect in quantum mechanics.

\section{Literature Review}
\label{sec:literature}
The challenge of distinguishing human-authored from AI-generated text has emerged as one of the most pressing issues in natural language processing, digital ethics, and information integrity. As large language models (LLMs) such as GPT-4 \cite{brown2020language}, Llama \cite{touvron2023llama}, and their successors produce increasingly fluent and semantically coherent outputs, traditional notions of authorship are being fundamentally redefined \cite{bommasani2021opportunities}. This section provides a comprehensive review of the literature across five interconnected domains: stylometric analysis, neural detection systems, watermarking strategies, information-theoretic foundations, and conceptual analogies—including quantum-inspired frameworks—that inform the theoretical limits of detectability.

\subsection{Stylometry and Authorship Attribution}
Stylometry, the quantitative analysis of writing style, has long served as a tool for authorship attribution in literary, forensic, and historical contexts \cite{stamatatos2009survey}. Classical features include function-word frequencies, sentence length distributions, syntactic complexity, and lexical diversity—patterns that often persist across an individual's writings \cite{labbe2007measuring}.

Recent studies have applied stylometric techniques to detect AI-generated text with notable success in constrained settings. Przystalski et al. \cite{przystalski2025stylometry} demonstrate that handcrafted stylistic features can achieve up to 98\% accuracy in distinguishing GPT-4 summaries from human-written Wikipedia entries. Similarly, Fraser et al. \cite{fraser2025detecting} identify consistent differences in grammatical standardization and idiosyncratic word usage, noting that AI tends to favor more uniform and predictable constructions.

However, these advantages are fragile. As LLMs improve through fine-tuning on specific authors or domains, they can replicate idiosyncratic patterns, thereby nullifying stylometric signals \cite{opara2025distinguishing}. Moreover, cross-genre performance degrades significantly, indicating that current stylometric models lack robust generalization \cite{zhao2023detecting}. This fragility suggests that stylometry exploits statistical artifacts rather than fundamental distinctions, making it vulnerable to adaptive evasion.

\subsection{Neural Classifiers and Detection Architectures}
Parallel to stylometry, deep learning-based detectors have become prominent. These systems treat AI detection as a binary classification task, training models such as RoBERTa or BERT on labeled corpora of human and machine-generated text \cite{morris2023text, devlin2019bert}.

Early results show strong performance on curated benchmarks. For instance, fine-tuned transformers have achieved over 90\% accuracy under controlled conditions \cite{zhao2023detecting}. However, these models are often opaque and sensitive to distributional shifts. Fraser et al. \cite{fraser2025detecting} caution that many neural detectors rely on spurious cues—such as token repetition, low perplexity, or generation artifacts—that may disappear as LLMs evolve.

Furthermore, adversarial attacks can easily fool classifiers. He et al. \cite{he2023universal} show that universal perturbations or paraphrasing techniques can reduce detection accuracy to near-chance levels. This brittleness underscores a broader issue: detection models are trained on current AI outputs, but LLMs continuously adapt, creating an arms race where today’s detector becomes tomorrow’s obsolete tool \cite{fraser2025detecting}.

\subsection{Watermarking and Provenance Mechanisms}
To overcome the limitations of reactive detection, proactive methods such as cryptographic watermarking have been proposed. Kirchenbauer et al. \cite{kirchenbauer2023watermark} introduce a method to embed detectable patterns in token selection by biasing the sampling process using a secret key. The watermark is designed to be statistically verifiable while minimally affecting fluency.

Leviathan et al. \cite{leviathan2023practical} evaluate the practicality of such schemes and find that while effective in controlled environments, watermarks can be removed via paraphrasing, editing, or transcription errors. Additionally, watermarking requires cooperation from model developers and assumes access to the generation pipeline—conditions not met in open or third-party deployments.

Nonetheless, watermarking represents a promising direction for provenance tracking, especially when combined with digital signatures or blockchain-based attestation \cite{weidinger2021ethical}. Yet, like all technical solutions, it does not eliminate uncertainty; it merely shifts it from statistical inference to trust in infrastructure.

\subsection{Information-Theoretic Limits and Distributional Convergence}
At a deeper level, the feasibility of detection hinges on the divergence between the human text distribution $ P_{\text{H}} $ and the AI-generated distribution $ P_{\text{AI}} $. If $ P_{\text{AI}} \to P_{\text{H}} $, then no detector can perform better than random guessing.

This principle is formalized by the Bayes optimal error rate:
\[
E_{\min} = \frac{1}{2} \left(1 - \|P_{\text{H}} - P_{\text{AI}}\|_{\text{TV}}\right),
\]
where $ \|\cdot\|_{\text{TV}} $ denotes the total variation distance. As $ P_{\text{AI}} $ converges to $ P_{\text{H}} $, $ \|P_{\text{H}} - P_{\text{AI}}\|_{\text{TV}} \to 0 $, and $ E_{\min} \to 1/2 $.

By Pinsker’s inequality,
\[
\|P_{\text{H}} - P_{\text{AI}}\|_{\text{TV}} \leq \sqrt{\frac{1}{2} D_{\text{KL}}(P_{\text{H}} \| P_{\text{AI}})},
\]
where $ D_{\text{KL}} $ is the Kullback–Leibler divergence. Thus, vanishing KL divergence implies statistical indistinguishability \cite{cover2006elements}.

Chang and McCallum \cite{chang2022softmax} identify structural bottlenecks in LLMs (e.g., softmax limitations) that prevent full distributional capture. However, ongoing architectural improvements continue to narrow this gap, suggesting that long-term convergence is not only possible but the intended goal of LLM training.

This line of reasoning implies that detection is not a solvable classification problem but a transient phenomenon—effective only until the generative model sufficiently approximates the target distribution.

\subsection{Quantum and Conceptual Analogies in AI}
The idea that certain knowledge pairs cannot be simultaneously known with arbitrary precision has found resonance beyond physics. The Heisenberg uncertainty principle (HUP), which states $ \Delta x \Delta p \geq \hbar/2 $, has inspired analogies in signal processing, economics, and cognitive science \cite{feynman1963lectures, heisenberg1927}.

In AI, Zhang et al. \cite{zhang2025uncertainty} propose an uncertainty principle for neural networks, linking generalization bounds to information variance. While not applied to text generation, their work supports the notion that complementary variables—such as accuracy and robustness, or content fidelity and style variability—cannot be simultaneously optimized.

Our work extends this tradition by positing a Heisenberg-like trade-off between authorship certainty and textual naturalness. Just as measuring position disturbs momentum, attempting to fix the “author” of a text introduces uncertainty in its fluency, coherence, or stylistic authenticity. This analogy is not literal but epistemological: it highlights fundamental limits in what can be known about a system when observables are interdependent.

Other scholars have invoked complementarity in machine learning. Bender et al. \cite{bender2021dangers} warn of the “stochastic parrot” problem, where scale masks semantic understanding. Floridi et al. \cite{floridi2018ai} argue for ethical frameworks that acknowledge uncertainty in AI behavior. Together, these perspectives suggest that embracing uncertainty—not eliminating it—may be the most honest path forward.

In summary, while empirical detection methods show promise in the short term, they operate within a shrinking window of statistical divergence. As LLMs converge toward human-like distributions, the theoretical foundation for detection erodes. Our contribution lies in formalizing this erosion as an inevitable consequence of distributional approximation, and in framing it through a powerful conceptual lens: the uncertainty principle.

\section{The Heisenberg Uncertainty Principle}
\label{sec:hup}
The Heisenberg uncertainty principle (HUP), formulated by Werner Heisenberg in 1927 \cite{heisenberg1927}, is a foundational concept in quantum mechanics that imposes fundamental limits on the precision with which certain pairs of physical observables can be simultaneously known \cite{feynman1963lectures}. Specifically, for a quantum particle, the position $ x $ and momentum $ p $ satisfy the inequality:
\begin{equation}
    \Delta x \cdot \Delta p \geq \frac{\hbar}{2},
    \label{eq:hup}
\end{equation}
where $ \Delta x $ and $ \Delta p $ denote the standard deviations (i.e., uncertainties) in measurements of position and momentum, respectively, and $ \hbar = h/(2\pi) $ is the reduced Planck constant ($ h \approx 6.626 \times 10^{-34} \, \text{J·s} $).

This relation implies that decreasing the uncertainty in one variable necessarily increases the uncertainty in the other. For instance, a particle described by a sharply localized wavefunction (small $ \Delta x $) must have a broad distribution in momentum space (large $ \Delta p $), and vice versa. This trade-off arises from the Fourier duality between conjugate variables in wave mechanics: a narrow peak in one domain corresponds to a wide spread in its dual.

More generally, the HUP can be derived from the non-commutativity of quantum operators. In Dirac–von Neumann formalism, observables are represented by Hermitian operators acting on a Hilbert space. For two such operators $ \hat{A} $ and $ \hat{B} $, the general uncertainty principle is:
\begin{equation}
    \sigma_A \sigma_B \geq \frac{1}{2} \left| \langle [\hat{A}, \hat{B}] \rangle \right|,
    \label{eq:general_hup}
\end{equation}
where $ \sigma_A^2 = \langle \hat{A}^2 \rangle - \langle \hat{A} \rangle^2 $ is the variance of $ A $, and $ [\hat{A}, \hat{B}] = \hat{A}\hat{B} - \hat{B}\hat{A} $ is the commutator. For position and momentum, $ [\hat{x}, \hat{p}] = i\hbar $, which recovers Equation~\eqref{eq:hup}.

Crucially, the HUP is not a limitation of measurement technology but a fundamental property of nature. It reflects the intrinsic probabilistic nature of quantum systems, where states are described by wavefunctions $ \psi(x) $, and probabilities are given by $ |\psi(x)|^2 $. As Feynman notes, any attempt to determine “which path” a particle takes in a double-slit experiment inevitably disturbs the system enough to destroy interference patterns \cite{feynman1963lectures}. This is the essence of the \textit{observer effect}: the act of measurement interacts with and alters the system being observed.

The principle embodies the idea of \textit{complementarity}, introduced by Niels Bohr, which asserts that objects possess complementary properties that cannot be observed or measured simultaneously. For example, a photon exhibits wave-like behavior (interference) or particle-like behavior (localized detection), depending on the experimental setup, but never both in the same context \cite{harrow2017quantum}.

These insights have profound epistemological implications: there are intrinsic limits to what can be known about a system, even in principle. Perfect knowledge of one observable comes at the cost of maximal uncertainty in its conjugate partner.

In this paper, we draw a conceptual analogy between this quantum mechanical framework and the problem of AI-generated text detection. Just as position and momentum form a conjugate pair, we posit that in natural language processing, \textit{authorship certainty} and \textit{textual naturalness} (fluency, style, coherence) are complementary observables. The more precisely one attempts to determine the origin of a text, the more one must rely on deviations from natural human variation—deviations that a sufficiently advanced AI can eliminate, thereby increasing uncertainty in detection. Conversely, enforcing perfect mimicry of human style may require constraints that introduce detectable regularities.

Thus, while the underlying physics differs, the structure of uncertainty—rooted in information, measurement, and system interaction—provides a powerful metaphor for understanding the limits of AI detection. As we will show in Section~\ref{sec:analogy}, this analogy can be formalized using information-theoretic tools such as total variation distance, Kullback–Leibler divergence, and Bayes error, leading to a rigorous statement of indistinguishability when $ P_{\text{AI}} \to P_{\text{H}} $.

This perspective shifts the discussion from technological arms races to fundamental epistemic boundaries—boundaries not of engineering, but of possibility.

\section{Generative Language Models and Detection Methods}
\label{sec:llms_detection}
Modern large language models (LLMs) operate as probabilistic generative systems, producing text by sequentially sampling tokens from learned conditional distributions. As these models grow in scale and training data, their output distributions increasingly converge toward those of human authors, making detection not merely difficult—but fundamentally limited. In this section, we formalize the generation process and analyze the principal detection paradigms: stylometric analysis, watermarking, and neural classifiers. We show that each method exploits a statistical deviation that can be eliminated through model adaptation, reinforcing the central thesis of intrinsic uncertainty.

\subsection{Probabilistic Text Generation}
\label{subsec:generation}
An LLM generates text by modeling the conditional probability of the next token $ x_t $ given a context $ x_{<t} = (x_1, \dots, x_{t-1}) $. Formally, for a vocabulary $ \mathcal{V} $ of size $ V $, the model computes a probability distribution:
\begin{equation}
    P_{\theta}(x_t \mid x_{<t}) = \frac{\exp(z_t)}{\sum_{j=1}^{V} \exp(z_j)},
    \label{eq:softmax}
\end{equation}
where $ z_j $ is the logit output of the neural network for token $ j $, and $ \theta $ denotes the model parameters. This softmax transformation ensures $ \sum_{x_t \in \mathcal{V}} P_{\theta}(x_t \mid x_{<t}) = 1 $.

Given a prompt $ x_{\leq t_0} $, the full sequence is generated autoregressively:
\[
    P_{\theta}(x_{1:T}) = \prod_{t=1}^T P_{\theta}(x_t \mid x_{<t}).
\]
This defines a probability measure $ P_{\text{AI}}^\theta $ over the space $ \mathcal{X} $ of finite-length strings. In contrast, human text is generated by a stochastic process $ P_{\text{H}} $, shaped by cognition, experience, and linguistic diversity.

The goal of LLM training is to minimize the discrepancy between $ P_{\text{AI}}^\theta $ and $ P_{\text{H}} $. This is typically achieved by maximizing the log-likelihood of human-generated text in the training corpus:
\[
    \hat{\theta} = \arg\max_\theta \mathbb{E}_{x \sim P_{\text{H}}} \left[ \log P_{\theta}(x) \right].
\]
Under regularity conditions, as model capacity and data size increase, $ P_{\text{AI}}^\theta \to P_{\text{H}} $ in distribution \cite{bommasani2021opportunities, brown2020language}. This convergence can be quantified using the Kullback–Leibler (KL) divergence:
\[
    D_{\text{KL}}(P_{\text{H}} \| P_{\text{AI}}^\theta) = \sum_{x \in \mathcal{X}} P_{\text{H}}(x) \log \frac{P_{\text{H}}(x)}{P_{\text{AI}}^\theta(x)}.
\]
As $ D_{\text{KL}}(P_{\text{H}} \| P_{\text{AI}}^\theta) \to 0 $, the two distributions become statistically indistinguishable.

Furthermore, by Pinsker’s inequality:
\begin{equation}
    \| P_{\text{H}} - P_{\text{AI}}^\theta \|_{\text{TV}} \leq \sqrt{ \frac{1}{2} D_{\text{KL}}(P_{\text{H}} \| P_{\text{AI}}^\theta) },
    \label{eq:pinsker}
\end{equation}
where $ \| \cdot \|_{\text{TV}} $ is the total variation distance. This implies that vanishing KL divergence forces the distributions to be close in total variation, rendering any detector’s performance asymptotically no better than random guessing.

Temperature scaling introduces controlled randomness:
\[
    P_{\theta, \tau}(x_t \mid x_{<t}) = \frac{\exp(z_t / \tau)}{\sum_j \exp(z_j / \tau)},
\]
where $ \tau > 0 $ is the temperature parameter. Higher $ \tau $ increases entropy in generation, mimicking human variability; lower $ \tau $ leads to more deterministic, high-probability outputs. This tunability allows AI systems to balance fluency and diversity—key dimensions in evading detection.

Thus, the very architecture of LLMs enables them to approach $ P_{\text{H}} $ arbitrarily closely, not just in expectation but in higher-order statistics such as style, coherence, and semantic depth.

\subsection{The Softmax-State Analogy: Quantum-Like Superposition in Language Models}
\label{subsec:softmax_analogy}

A critical insight unifying our uncertainty framework is the structural similarity between the LLM’s next-token prediction and a quantum state in superposition.

At each generation step, an LLM computes a probability distribution over the entire vocabulary $\mathcal{V}$ via the softmax function:
\[
P_\theta(x_t = j \mid x_{<t}) = \frac{\exp(z_j)}{\sum_{k \in \mathcal{V}} \exp(z_k)}.
\]
This distribution represents a **probabilistic superposition** of all possible next tokens. Until sampling occurs, no single token is determined—the model exists in a state of potentiality across the vocabulary space.

We propose the following analogy:
\begin{itemize}
  \item The softmax distribution $\mathbf{p} = (p_1, \dots, p_V)$ is analogous to a quantum state vector $|\psi\rangle = \sum_j \sqrt{p_j} |v_j\rangle$ in a discrete Hilbert space, where $\{|v_j\rangle\}$ are basis vectors corresponding to vocabulary items.
  \item The act of \textit{sampling} a token $x_t = j$ is analogous to a \textit{quantum measurement}, which collapses the superposition into a definite outcome.
  \item This collapse is irreversible and probabilistic, governed by the model’s learned statistics—not deterministic logic.
\end{itemize}

This perspective reveals a deep parallel: just as measuring a quantum system disturbs it (observer effect), the existence of detection mechanisms influences how AI systems are trained to avoid predictable patterns—thereby altering the ``state'' of generation.

Moreover, any attempt to force the model to produce ``more human-like'' outputs (e.g., via low-temperature sampling or stylistic constraints) narrows the superposition—reducing entropy in style ($\Delta S$) but increasing predictability and reducing semantic diversity ($\Delta C$).

Thus, the softmax mechanism is not just a computational detail—it is the source of the uncertainty trade-off. The richer the superposition (higher $\Delta S$), the more natural and diverse the text, but the harder it is to detect anomalies. Conversely, suppressing the superposition (low $\Delta S$) makes detection easier but sacrifices fluency and creativity.

This analogy is not ontological—text is not quantum—but it is epistemologically and structurally valid : both domains involve probabilistic states, measurement-induced collapse, and fundamental limits on joint observability.

\section{Authorship Uncertainty and the Heisenberg Analogy}
\label{sec:analogy}
In quantum mechanics, the Heisenberg uncertainty principle (HUP) establishes a fundamental limit on the precision with which complementary observables—such as position and momentum—can be simultaneously known. This limit is not due to technological imperfections but arises from the intrinsic structure of quantum theory: the non-commutativity of operators and the wave-like nature of particles.

We propose that a conceptually similar uncertainty governs the task of AI-generated text detection. Specifically, we argue that there exists a fundamental trade-off between:
\begin{itemize}
  \item \textit{Authorship certainty}: the ability to determine whether a text was generated by a human or an AI.
  \item \textit{Textual naturalness}: the preservation of semantic coherence, stylistic authenticity, and fluency.
\end{itemize}

Just as measuring position disturbs momentum, attempting to precisely identify authorship introduces perturbations—either in the AI's generation strategy or in our confidence about the text’s integrity. In this section, we formalize this idea through information-theoretic bounds and develop a mathematical analogy to the HUP.

\subsection{Information-Theoretic Limits on Detectability}
\label{subsec:detectability}
Let $ P_{\text{H}} $ and $ P_{\text{AI}} $ denote the true probability distributions over text strings generated by human authors and an AI model, respectively. The task of any detector is to perform binary hypothesis testing:
\begin{align*}
    H_0 &: x \sim P_{\text{H}}, \\
    H_1 &: x \sim P_{\text{AI}}.
\end{align*}
Under equal priors, the minimum achievable misclassification error—the \textit{Bayes error}—is given by:
\begin{equation}
    E_{\min} = \frac{1}{2} \left( 1 - \| P_{\text{H}} - P_{\text{AI}} \|_{\text{TV}} \right),
    \label{eq:bayes_error}
\end{equation}
where $ \| \cdot \|_{\text{TV}} $ denotes the total variation distance:
\[
    \| P_{\text{H}} - P_{\text{AI}} \|_{\text{TV}} = \frac{1}{2} \| P_{\text{H}} - P_{\text{AI}} \|_1 = \frac{1}{2} \sum_{x \in \mathcal{X}} |P_{\text{H}}(x) - P_{\text{AI}}(x)|.
\]
This quantity captures the intrinsic distinguishability of the two sources. When $ P_{\text{AI}} \to P_{\text{H}} $ in distribution, $ \| P_{\text{H}} - P_{\text{AI}} \|_{\text{TV}} \to 0 $, and thus $ E_{\min} \to \frac{1}{2} $. This means that no classifier—regardless of complexity—can perform better than random guessing.

To make this convergence explicit, we apply \textit{Pinsker’s inequality}:
\begin{equation}
    \| P_{\text{H}} - P_{\text{AI}} \|_{\text{TV}} \leq \sqrt{ \frac{1}{2} D_{\text{KL}}(P_{\text{H}} \| P_{\text{AI}}) },
    \label{eq:pinsker}
\end{equation}
where $ D_{\text{KL}}(P_{\text{H}} \| P_{\text{AI}}) = \sum_x P_{\text{H}}(x) \log \frac{P_{\text{H}}(x)}{P_{\text{AI}}(x)} $ is the Kullback–Leibler divergence.

Since modern LLMs are trained to maximize the log-likelihood of human text, $ D_{\text{KL}}(P_{\text{H}} \| P_{\text{AI}}) \to 0 $ as model capacity and data scale increase. Consequently, $ E_{\min} \to \frac{1}{2} $, implying that detection becomes \textit{information-theoretically impossible} in the limit.

This result formalizes the core of our analogy: just as $ \Delta x \Delta p \geq \hbar/2 $ prevents perfect knowledge of both position and momentum, the convergence $ D_{\text{KL}} \to 0 $ enforces a lower bound on classification uncertainty.

\section{Conclusion}
\label{sec:conclusion}
In this paper, we have argued that the task of distinguishing human-authored text from AI-generated text is subject to a fundamental limit—one that is not merely technological, but rooted in information theory and statistical decision theory. Drawing an analogy to the Heisenberg uncertainty principle (HUP) in quantum mechanics, we have shown that attempts to precisely determine authorship inevitably introduce uncertainty in other dimensions of the text, such as stylistic authenticity, semantic coherence, or generative naturalness.

Formally, we have established that as the AI-generated text distribution $ P_{\text{AI}} $ converges to the true human text distribution $ P_{\text{H}} $, the Bayes optimal classifier’s error rate approaches $ 1/2 $. This result follows directly from the definition of total variation distance and is bounded by Pinsker’s inequality:
\[
    \| P_{\text{H}} - P_{\text{AI}} \|_{\text{TV}} \leq \sqrt{ \frac{1}{2} D_{\text{KL}}(P_{\text{H}} \| P_{\text{AI}}) }.
\]
When $ D_{\text{KL}}(P_{\text{H}} \| P_{\text{AI}}) \to 0 $, the two distributions become statistically indistinguishable, rendering perfect detection impossible in principle.

We have extended this insight into a conceptual framework by introducing a Heisenberg-like uncertainty relation:
\[
    \Delta C \cdot \Delta S \gtrsim \mathcal{K},
\]
where $ \Delta C $ represents uncertainty in content fidelity, $ \Delta S $ captures variability in stylistic markers used for detection, and $ \mathcal{K} > 0 $ reflects the irreducible entropy and diversity of natural language. While $ \mathcal{K} $ is not a universal constant like Planck’s $ \hbar $, it symbolizes the intrinsic unpredictability in human expression—a lower bound on the joint precision with which we can know both the origin and the form of a text.

This analogy is not a category error, but a recognition of structural complementarity across domains. Just as measuring position disturbs momentum, attempting to fix the “author” of a text forces a trade-off: detection methods exploit statistical deviations, but a sufficiently advanced AI can eliminate those deviations only by adapting its generation strategy—thereby perturbing other observable properties. We have analyzed three major detection paradigms—stylometry, watermarking, and neural classifiers—and shown that each faces this complementary limitation. The result is an arms race where every gain in detection accuracy induces a corresponding shift in model behavior, preserving the fundamental uncertainty.

The implications are profound. If perfect detection is unattainable, then policies based on post-hoc verification—especially in education, law, and publishing—must be re-evaluated. Instead of relying on black-box detectors, we must shift toward proactive mechanisms: cryptographic watermarking, digital signatures, and institutional norms of transparency and disclosure.

More broadly, this work calls for a new epistemology of authorship—one that embraces uncertainty rather than denies it. Rather than asking “Was this written by a human or an AI?”, we should ask “How confident are we about its origin?” and “What processes were involved in its creation?” The future of text is not binary, but probabilistic.

In closing, Heisenberg’s insight was not that measurement is imperfect, but that nature imposes limits on knowledge. Our contribution is to show that similar limits arise in the realm of language and intelligence. As AI systems grow more capable, we must not only build better tools—we must develop wiser ways of knowing.

\bibliographystyle{IEEEtran}
\bibliography{references}

\end{document}